\documentclass[lettersize,journal]{IEEEtran}
\usepackage{amsmath,amsfonts}
\usepackage{algorithmic}
\usepackage{algorithm}
\usepackage{array}
\usepackage[caption=false,font=normalsize,labelfont=sf,textfont=sf]{subfig}
\usepackage{textcomp}
\usepackage{stfloats}
\usepackage{url}
\usepackage{verbatim}
\usepackage{graphicx}
\usepackage{cite}

\usepackage{makecell}
\usepackage{multirow}
\usepackage{multicol}
\usepackage{booktabs}
\usepackage{colortbl}
\usepackage{arydshln}
\usepackage{xcolor}

\IEEEoverridecommandlockouts

\IEEEpubid{\begin{minipage}{\textwidth}\ \\[12pt] \centering
  Copyright © 2023 IEEE. Personal use of this material is permitted.
  However, permission to use this material for any other purposes must be
  obtained 
  from the IEEE by sending an email to pubs-permissions@ieee.org.
\end{minipage}}

\begin{document}

\title{Self-Training Vision Language BERTs with a Unified Conditional Model}

\author{Xiaofeng Yang, Fengmao Lv, Fayao Liu, Guosheng Lin 
\thanks{Corresponding author: Guosheng Lin.}
\thanks{Xiaofeng Yang and Guosheng Lin are with School of Computer Science and Engineering, Nanyang Technological University (NTU), Singapore 639798 (email: xiaofeng001@e.ntu.edu.sg, gslin@ntu.edu.sg)}

\thanks{Fengmao Lv is with School of Computing and Artificial Intelligence, Southwest Jiaotong University, Chengdu 611756, China (email: fengmaolv@126.com)}

\thanks{Fayao Liu is with Agency for Science, Technology and Research (A*STAR), Singapore 138632 (email: fayaoliu@gmail.com)}

}
\markboth{Journal of \LaTeX\ Class Files,~Vol.~14, No.~8, August~2021}%
{Shell \MakeLowercase{\textit{et al.}}: A Sample Article Using IEEEtran.cls for IEEE Journals}

\maketitle
\IEEEpubidadjcol

\begin{abstract}
Natural language BERTs are trained with language corpus in a self-supervised manner. Unlike natural language BERTs, vision language BERTs need paired data to train, which restricts the scale of VL-BERT pretraining. We propose a self-training approach that allows training VL-BERTs from unlabeled image data. The proposed method starts with our unified conditional model -- a vision language BERT model that can perform zero-shot conditional generation. Given different conditions, the unified conditional model can generate captions, dense captions, and even questions. We use the labeled image data to train a teacher model and use the trained model to generate pseudo captions on unlabeled image data. We then combine the labeled data and pseudo labeled data to train a student model. The process is iterated by putting the student model as a new teacher. By using the proposed self-training approach and only 300k unlabeled extra data, we are able to get competitive or even better performances compared to the models of similar model size trained with 3 million extra image data.
\end{abstract}


\section{Introduction}
Large scale pretraining has become the dominating approach in various natural language processing tasks. The success of large scale pretraining is due to a large amount of language training data available everywhere and the self-training algorithm. Unlike language pretraining, vision language pretraining requires paired image and language data, which restricts the scale of vision language BERTs' pretraining. In this paper, we propose a self-training approach that allows to pretrain VL-BERTs using unlabeled image data. 

Self-training is usually done by iterating the following three steps: 1) training with labeled data, 2) generating pseudo labels for unlabeled data, 3) mixing the labeled data and unlabeled data with pseudo labels to retrain the network. However, the self-training of vision language BERTs is  nontrivial due to the following reasons. First, although auto-encoding models (e.g.,  BERTs~\cite{devlin2018bert,lu2019vilbert}) perform well on the natural language understanding and image language understanding tasks,  they cannot be directly applied to the generation task without finetuning~\cite{wang2019bert}. In practice, it is difficult to generate pseudo labels for unlabeled data using pretrained BERTs in the zero-shot setting. Although these models can be finetuned to perform generation tasks, the zero-shot generation of pseudo labels is important since it saves the time of extra finetuning and avoids adding additional bias from the finetuning datasets. Second, current common practice in vision language BERT pretraining uses various image descriptions to train, such as image captions, dense captions and questions. Those image descriptions have significant differences, making it difficult for an unconditional model to learn to generate adequate pseudo captions for unlabeled images. Hence, although self-training has shown its effectiveness in various tasks~\cite{xie2020self,zoph2020rethinking}, how to use it effectively in training vision language BERTs is not yet studied. 

To this end, we propose the Unified Conditional Model (UCM) and a set of vision language BERT self-training methods to tackle the above issues. Compared with previous methods, our model has the following advantages: First, our method combines auto-encoding training~\cite{devlin2018bert,lu2019vilbert} and auto-regressive training~\cite{brown2020language} in a unified framework, which enables our method to perform well on natural language understanding tasks and at the same time effectively generate pseudo labels. Second, we propose a novel conditional training method that enables our model to conditional generate various types of captions, including COCO style captions, VG style dense captions and questions.

\begin{figure}[]
\centering
\includegraphics[width=0.5\textwidth,trim=0 0 0 0,clip]{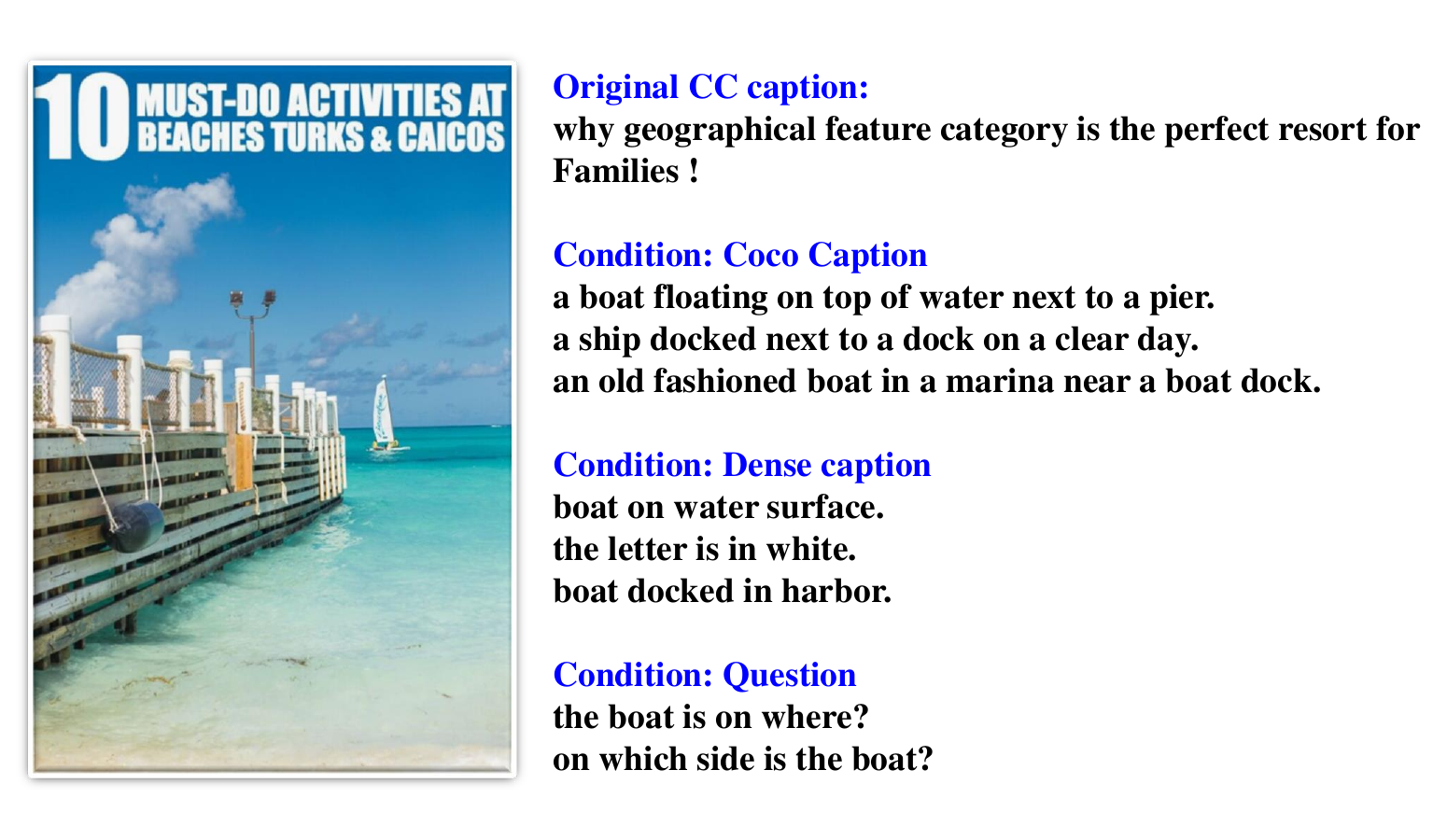}
\caption{An example of generated image descriptions. The original image is selected from Conceptual Caption. Given different condition flags, our proposed UCM model is able to generate diverse image descriptions, such as COCO caption, dense caption, and questions. It's clear that the generated contents have different styles. Compared with the originally provided captions, the generated ones could better describe the picture contents.}
\label{fig:1}
\end{figure}
\IEEEpubidadjcol 
\textbf{Unified Conditional Model (UCM).} Compared with traditional vision language BERTs, our proposed UCM has two unique properties. First, the model is able to generate different contents based on a condition flag input, such as image captions, dense captions, and questions. Second, the condition flag can be used as an identifier to help down-stream finetuning. The proposed UCM shares similar model structures with existing 2-stream vision language BERT models~\cite{tan2019lxmert,lu2019vilbert}. Specifically, it contains one image encoder, two language encoders with shared weights, and several layers of cross attention layers. In training, different data types are assigned to their condition signals. The model is trained with both bi-directional prediction masks and one-directional prediction masks in parallel. For bi-directional prediction masks, the model performs conditional masked language modeling prediction, masked object prediction, image-text matching, and an auxiliary question answering loss. For one-directional prediction masks, the model performs one-directional masked conditional language modeling and masked object prediction tasks. When the model is used to generate pseudo labels for unlabeled images, the model will run forward propagation with one-directional prediction masks only. The condition signal enables the model to generate diverse descriptions for pictures. Fig.~\ref{fig:1} shows an example of generated image descriptions using different condition flags. When the model is used for finetuning image language understanding tasks, only the bi-directional mask is used. During finetuning, we use the condition flag as prior knowledge for finetuning. For example, when finetuning VQA tasks, the input is given an additional condition flag to show the input is a question. Results show that the presence of condition flags improves down-stream finetuning performance. 

\textbf{Vision Language BERT Self-Training.} The self-training method is used in the pretraining stage to further enlarge the scale of data that can be used in pretraining. Our self-training approach follows the self-training pipeline with extra optimization for vision language BERTs. Generally, the self-training process is done in three steps. First, we use the labeled image data to train a teacher model and then use the trained model to generate pseudo labels on unlabeled image data.  We then combine the labeled data and pseudo labeled data to train a student model. Finally, the process is iterated by putting the student model as a new teacher. In our task, the pseudo labels are generated COCO style image captions and VG style dense captions by UCM. In order to generate high quality and diverse pseudo labels, we propose three methods. First, we randomly mask object regions when generating captions. This method makes sure the model can focus on different visual areas when describing the images. Second, we randomly sample a word from the top-K predictions in each prediction step, such that even for the same image, the model could generate various outputs. Finally, we use the condition flag to control the contents generated. 
We show both qualitative and quantitative comparisons in experiments section.

Experimental-wise, besides the commonly used COCO and VG datasets, we train our model with only 300k extra unlabeled data from Conceptual Caption~\cite{sharma2018conceptual} by removing the provided captions. The original Conceptual Caption dataset provides machine-generated captions. They are noisy~\cite{singh2020we} and often used as out-of-domain training data~\cite{chen2019uniter}. The model could out-perform the model trained with the whole three million extra data in various down-stream finetuning tasks. Also, we provide comprehensive ablation studies of the training settings.  To summarize our contributions:
\begin{itemize}
  \item We propose the first Unified Conditional BERT model that could perform zero-shot conditional image-based language generation. Traditional bi-directional vision language models are unable to be used to generate languages directly and they are not conditional, such that the users can't control the generation styles.
  \item We propose a self-training method for using unlabeled images in vision language pretraining. To the best of our knowledge, this is the first work using self-training in vision language pretraining. 
  \item With only 300k extra image data, we achieve competitive or better performances within models with similar model size trained with 3 million extra data.
\end{itemize}

\section{Related Work}
\subsection{Vision Language Pretraining}
Traditional vision language methods build stand-alone models to solve VQA~\cite{anderson2018bottom,han2019movie,zhang2020action,guo2021loss,wang2017fvqa,guo2021re}, captioning~\cite{yu2018topic,huang2020image,yan2021task}, navigation~\cite{zhang2020language} and grounding~\cite{gao2022efficient} tasks. The success of large scale pretraining in NLP~\cite{devlin2018bert} motivates the attempts of developing similar models in vision language. Original pretrained language models~\cite{devlin2018bert} use a single transformer~\cite{vaswani2017attention} to encode language words and positions. In the situations of vision + language, there are usually two common choices: the one-stream methods and two-stream methods. Two-stream methods, for example ViLBERT~\cite{lu2019vilbert}, LXMERT~\cite{tan2019lxmert} and 12in1~\cite{lu202012}, use two transformers to encode images and languages separately. After that, there will usually be a cross-attention transformer to combine the features from the two branches. One-stream methods, for example VisualBERT~\cite{li2019visualbert}, Unicoder-VL~\cite{li2020unicoder}
, Uniter~\cite{chen2019uniter} and Oscar~\cite{li2020oscar}, process vision and language features with a single transformer encoder. In this case, the visual and language information share the same attention weights. Compared with the two-stream methods, the one-stream methods require more working memories and usually perform better than two-stream methods. The one-stream methods usually have a smaller model size. Our work follows the two-stream network design of LXMERT~\cite{tan2019lxmert} and extends the single language encoder of LXMERT~\cite{tan2019lxmert} to two shared-weight language encoders that process the one-directional mask and two-directional mask at the same time. This network design allows our network to generalize better on generation tasks. 

Although BERT is a form of language model, same as natural language BERTs~\cite{devlin2018bert,wang2019bert}, the above vision language BERTs can not be used directly to generate languages. The most straightforward reason is that BERT learns bidirectional contexts, while generation is one-directional. VLP~\cite{zhou2020unified} proposes to train vision language models with both bi-directional and one-directional masks, such that the model can be used for both VQA and image captioning. Compared to previous work, our model has two unique properties: First, it is able to perform conditional generation, namely generating specific contents based on a condition signal. Second, we use the pretrained model to perform zero-shot language generation, without extra finetuning.

\subsection{Self-Training}
Self-training methods~\cite{yalniz2019billion,xie2020self} first use labeled data to train a teacher model, then use the teacher model to generate pseudo labels for unlabeled data and finally use the labeled data and pseudo labeled data to jointly train a student model. ~\cite{xie2020self} identifies the importance of add noise in self-training of image classification tasks. Self-training also improves object detection and semantic segmentation results compared with pre-training~\cite{zoph2020rethinking}. In machine translation~\cite{sennrich2015improving,cheng2019semi,wu2019exploiting}, self-trainings show their effectiveness on various datasets.

We provide a set of self-training algorithms and give detailed ablation studies.
\begin{figure*}[t]
\begin{center}
\includegraphics[width=0.9\linewidth,height=0.45\textwidth]{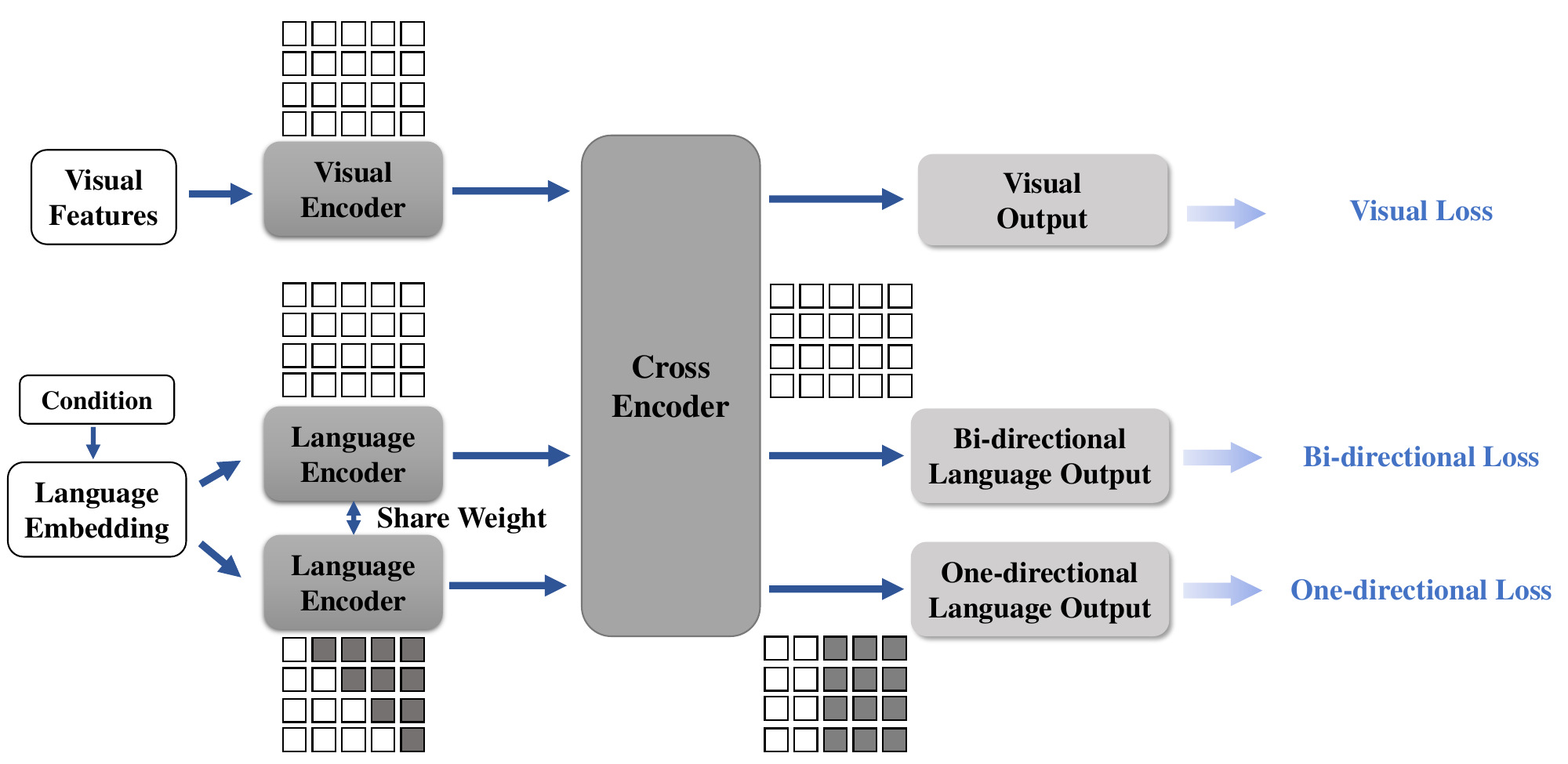}
	\caption{A detailed illustration of proposed UCM. During training, given image regional features and language embeddings, we process the language embeddings through the bi-directional language encoder and the one-directional language encoder. The two language encoders share the same weights. The bi-direction and one-directional branches are conditioned by using a normal mask and a triangular mask. The images are processed by the image encoder. Finally, the cross-attention layers merge visual features with the outputs from both language encoders. We use a rectangle mask for one-directional prediction in cross-attention layers, such that only the positions before [MASK] token could see visual features.}
	\label{fig:2}
\end{center}
\end{figure*}
\section{Method}
In this section, we describe our method in two folds: our proposed unified conditional model and the self-training algorithms. For the unified conditional model subsection, we first introduce the model overview including model structures and important tokens. Then we introduce the training tasks and training losses. For the self-training algorithms subsection, we introduce the technical details of our proposed self-training algorithms for vision language models. 

\subsection{Unified Conditional Model (UCM)}

\subsubsection{Model Overview} The overall structure of our model is illustrated in Figure~\ref{fig:2}. For the base network, we briefly follow the 2-stream model as used in~\cite{tan2019lxmert} and extend it to our unified conditional model. Specifically, the model contains 9 layers of language transformers, 5 layers of object region transformers, and 5 layers of cross transformers.  Given a sentence description, we tokenize it into WordPiece tokens, pad them with Classification [CLS] and Separation [SEP] tokens and randomly mask them with Mask [MASK] tokens. We add a condition token [CND] after the [CLS] token. The masked tokens are then passed to an embedding layer. We process the language embeddings through the bi-directional language encoder and the one-directional language encoder, same as previous works~\cite{dong2019unified,zhou2020unified}. The two language encoders share the same weights. The bi-direction and one-directional branches are distinguished by using an empty mask (bi-directional mask) and a triangular mask (one-directional mask)~\cite{radford2019language}. Given a one-directional mask, the tokens can only observe the tokens before themselves in the attention modules, which makes the module more capable of doing generation tasks. Given a bi-directional mask, the tokens can observe both the tokens after and before themselves. Experiments in BERT~\cite{devlin2018bert} prove that this design works better on understanding tasks, for example VQA tasks. The images are processed by the image encoder. After that, the bi-directional output and one-directional output are merged with image output through cross-attention layers~\cite{lu2019vilbert,tan2019lxmert}. For the cross-attention layer, we use a rectangle mask for the one-directional prediction branch, such that only the position before [MASK] could attend to image features.  During inference, our model is identical to traditional 2-stream vision language BERT models without any extra computational cost. When doing image-language understanding tasks, for example finetuning visual question answering, the model runs forward propagate using the bi-directional mask only. When performing image-language generation tasks, for example generating pseudo labeling for unannotated images, the model runs forward propagate using the one-directional mask only. 

\subsubsection{Training Tasks} 

\textbf{Conditional Masked Language Modeling (CMLM)} 
The CMLM task is for both bi-directional prediction and one-directional prediction. Given the image regional features $v = \{v_1, ..., v_K\}$, the language words $w =\{ w_1, ..., w_T \}$ and the condition $c$, for bi-directional prediction task, we randomly mask the language words at a ratio of 0.15. Once the position is masked, for 80\%, similar to BERT~\cite{devlin2018bert}, we change the position to [MASK] token and for 10\% of chance, we change to position to random word and keep the original content. The loss of bi-directional CMLM is defined as the negative log-likelihood of predicting the masked words given conditions and all other words except the mask words:

\begin{equation}
    \mathcal{L}_{\text{CMLM-Bi}}(\theta) = -\mathbb{E}_{(\mathbf{w}, \mathbf{v})} \log P_{\theta}(\mathbf{w}_\mathbf{m} | \mathbf{w}_{/ \mathbf{m}}, \mathbf{v},\mathbf{c})\,.
\end{equation}

For one-directional CMLM task, we randomly mask 1 word from each sentence. The masked word could also be the period symbol. The prediction of masked word is based on the words before the current position:

\begin{equation}
    \mathcal{L}_{\text{CMLM-One}}(\theta) = -\mathbb{E}_{(\mathbf{w}, \mathbf{v})} \log P_{\theta}(\mathbf{w}_\mathbf{m} | \mathbf{w}_{< \mathbf{m}}, \mathbf{v},\mathbf{c})\,,
\end{equation}

where $\theta$s in the above two equations represent the model parameters. Symbol $<$ represent all words before position m. The Bi-directional CMLM and the one-directional CMLM share the same model parameters.

 \textbf{Image-Text Matching} 
The matching task is only done for the bi-directional prediction branch. At 50\% possibility, we assign a fake sentence to the image. The fake sentence is generated by randomly sampling a caption from other images. Specifically, we use the final feature at position [CLS] to represent the summary of current visual and language input. We use this feature to classify whether the current input text and image are matched.

 \textbf{Auxiliary Question Answering} 
The QA task is only done for the bi-directional prediction branch. If the sampled text is a question, we use the feature at position [CLS] and a QA header to generate its answer and calculate its loss based on classification.

 \textbf{Masked Object and Attributes Modeling (MOAM)} 
The MOAM task is for both bi-directional prediction branch and one-directional prediction branch. The object features are always bidirectional visible for both branches. We randomly mask the visual regions at a ratio of 0.15. Once the region is masked, for 80\%, we change the feature to zero and for 10\% of chance, we change the feature to a random feature sampled from the dataset or keep the original feature. The loss of MOAM is defined as the negative log-likelihood of predicting the masked regions' class and attributes given all words except the masked position:

\begin{equation}
    \mathcal{L}_{\text{MOAM}}(\theta) = -\mathbb{E}_{(\mathbf{v}, \mathbf{w})} \log P_{\theta}(\mathbf{v}_\mathbf{m} | \mathbf{v}_{/ \mathbf{m}}, \mathbf{w})\,.
\end{equation}
Here, the ground-truth regional classes and attribute classes are hard labels generated by Faster-RCNN~\cite{anderson2018bottom} prediction. The MOAM losses from the two prediction branches are averaged when calculating gradients.

\textbf{Masked Feature Regression (MFR)} 
For each masked region, besides predicting the labels and attributes of that region, we also perform masked feature regression to recover its original visual feature:

\begin{equation}
    \mathcal{L}_{\text{MFR}}(\theta) = \|\mathbf{v}_\mathbf{m} - \mathbf{\hat{v}}_\mathbf{m} \|_2^2\,,
\end{equation}
where $\mathbf{\hat{v}}$ are the groundtruth regional features.
\begin{figure}[]
\centering
\includegraphics[width=0.45\textwidth,trim=0 0 0 0,clip]{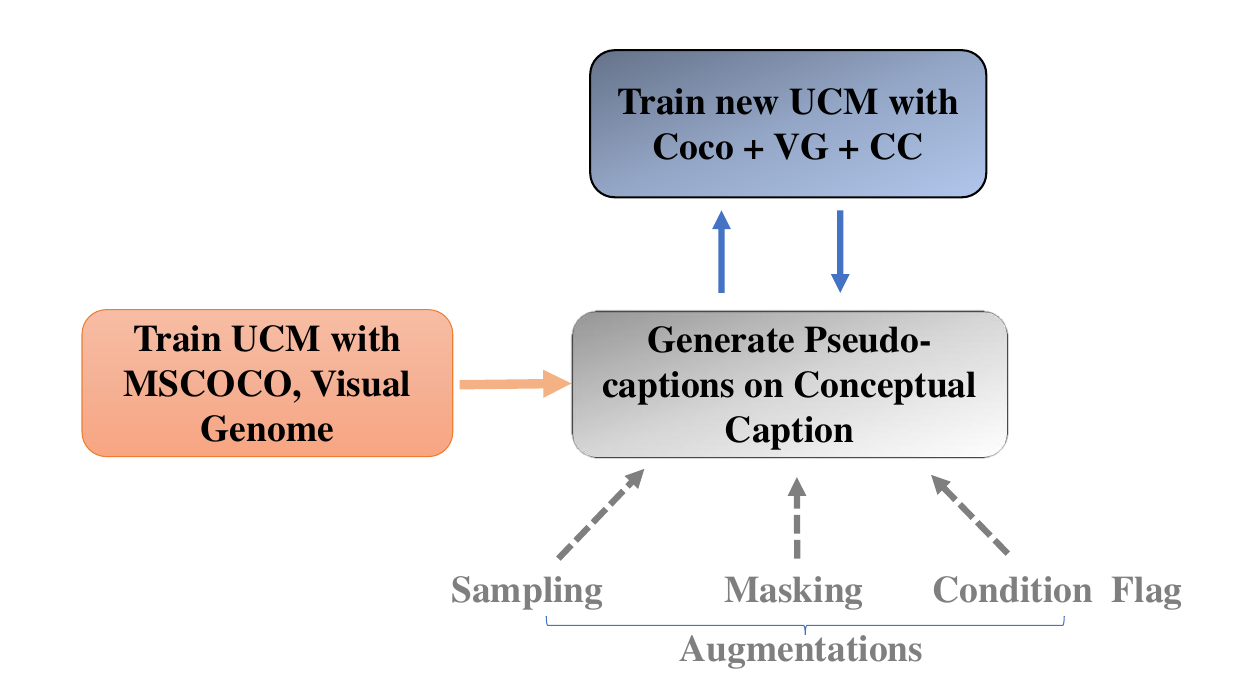}
\caption{The self-training algorithm. Our self-training approach is done by first training a UCM model with the labeled annotations and then iterating two steps: generating pseudo labeling on unlabeled data and retraining with mixed data. When generating pseudo labels, randomly sampling of language words, randomly masking image regions and condition flag are used as data augmentations.}
\label{fig:4}
\end{figure}

\subsection{Self-Training Algorithms for Vision Language BERTs}
In this section, we talk about the self-training algorithm. Figure~\ref{fig:4} illustrates the training process of our algorithm. We first train a UCM model with the human labeled data from COCO and VG datasets. Then we repeat two steps: generating pseudo labeling on Conceptual Caption unlabeled data and retraining the UCM model with mixed data from COCO, VG and Conceptual Caption.

\textbf{Train UCM with labeled data.} We first train UCM using captioning annotations in MSCOCO, dense captioning annotations in Visual Genome and questions in VQA~\cite{goyal2017making} and GQA dataset~\cite{hudson2019gqa}. The trained UCM is able to generate the above three different styles of content. 

\textbf{Annotate unlabeled data with trained UCM.} We then use the trained UCM to generate pseudo labels on images from Conceptual Captions dataset. Conceptual Caption dataset provides one caption for each image by default, while the default captions are machine-generated, not of good quality~\cite{singh2020we} and are often used as out-of-domain training data~\cite{chen2019uniter}. Therefore, we remove the original captions and use the data as unlabeled image data. To boost the performance of self-training, the generated captions need to be diverse. We introduce 3 methods to generate diverse image captions for each image. First, we perform image-level augmentations. We randomly mask object regions when generating captions. Empirically, each image regional feature is masked with a ratio of 0.5. This method makes sure the model can focus on different visual areas when generating outputs.  Second, we perform augmentations when sampling language words. We randomly sample a word from the top-K predictions in each prediction step, such that even for the same image input, the model could generate different captioning outputs. We choose $K=5$ based on the common design choice of Image Captioning~\cite{anderson2018bottom}. This is especially useful when generating dense captions. The result shows that the generated dense captions usually focus on one object region. Given one fixed image, the generated dense captions will always be the same without sampling. Compared with beam-search method, the top-K sampling method is faster but may potentially generate noisier captions. In experiments, we observe the performance differences between the two methods are negligible according to finetuning accuracy. One possible reason is that the generated captions are only used as pseudo labels and pseudo labels are noisy labels by default. Some works\cite{xie2020self} even purposely add noise to the generation process of pseudo labels. Therefore, we use top-K sampling method to speed up the generation process. We discard the generated questions because they usually contain less information than the captions. Finally, we use the condition flag to control the contents generated. For each image, we generate 5 captions with MSCOCO flag and 10 captions with VG Dense caption flag. The condition flag conditions the style of the generated contents.

\textbf{Train new model by mixing labeled and unlabeled data.} After pseudo labels are generated for unlabeled images, we mix the pseudo labeled data and original labeled data to train a new model. Unlike self-training methods in image classification~\cite{xie2020self}, which train new models from scratch, we propose to initialize the model with the last pretrained weight. The design choice is based on 2 considerations. First, vision language BERT pretraining takes a long time. Loading pretrained weight can help to reduce the training time of new models. Second, for image classification tasks, if we directly use soft classification labels to describe the unlabeled image and load previously trained weights, the loss will be zero on pseudo labeled data because the labels are exactly from the model itself. Compared with soft classification labels, generated captions are generated from sampling and do not directly describe the output distribution of previous models. This property reduces the risk that loading the previous model will result in zero loss on pseudo labeled data. This design choice also shares the same spirit as previous self-training works~\cite{tarvainen2017mean,athiwaratkun2018improving}, where the teacher models' weights are derived from the student models.

\textbf{Iterating the previous steps.} Following the common practice of self-training, we iterate the ``Annotate unlabeled data with trained UCM'' and ``Train new model by mixing labeled and unlabeled data'' steps a few times to get better performance. A detailed ablation study is shown in Section ~\ref{experiments}.

\section{Experiments}
\label{experiments}

In this section, we describe the pretraining details, ablation experiments, visualizations, and experimental results on down-stream datasets.
\subsection{Data Input} We first use pretrained Faster R-CNN~\cite{ren2015faster,anderson2018bottom} to extract regional visual features. The spatial information of each regional object is represented by its relative position, height, and weight. The spatial embedding is then calculated with an embedding layer. The final representation of each object is represented by adding the spatial embedding and visual features. For languages, we follow BERT~\cite{devlin2018bert} and tokenize the input sentences into WordPiece tokens. After tokenizing, We pad them with [CLS] and [SEP] tokens. Unlike original BERT, here we use [CLS] token to denote both start of sentence and classification position and we use [SEP] to denote end of sentence. Finally, we add the condition flag [CND] after [CLS] token. The condition flag [CND] represents a set of certain flags. In this work, the [CND] flag has three types: COCO type caption~\cite{chen2015microsoft}, visual genome~\cite{krishna2017visual} type dense caption and questions. 

\subsection{Pretrain Details}
\textbf{Pretraining Datasets.} Our pretraining dataset contains labeled data and unlabeled image data. For labeled data, we follow the same setting as in~\cite{tan2019lxmert}. The labeled data is collected from MSCOCO~\cite{chen2015microsoft}, Visual Genome~\cite{krishna2017visual}, VQA~\cite{goyal2017making} and GQA datasets~\cite{hudson2019gqa}, which contain around 180k images in total. Although the VG dataset contains more than 5 million image-text pairs, most of them are dense captions and some of them are repeated entries. In experiments, we remove the repeated dense captions, and sample 10 dense captions for each image. 
For unlabeled images, we use the first 300k images from Conceptual Caption dataset and remove the original captions. Within the 300k unlabeled images, we further filter the data by object detection results. We remove the images with the top 36 objects' average confidence below 0.3. Thus only 280k unlabeled images are left and we use them in self-training. The total numbers of pretraining images are illustrated in Table~\ref{nopairs}. Compared with LXMERT~\cite{tan2019lxmert} who uses 9 million image language pairs, we use a total amount of 7 million pairs. 

\begin{table}[]
\centering
\caption{Total Number of pretraining image-language pairs}
\begin{tabular}{cccccc}
\hline
                  & COCO & VG   &VQA &GQA & CC \\ \hline
Total \# of pairs & 533k & 5.06m &444k&1m& -  \\
Used \# of pairs   & 533k & 1m  &444k&1m& 4m \\ \hline
\end{tabular}
\label{nopairs}
\end{table}

\textbf{Self Training Setting.} In experiments, we iterate the self-training process 2 times. When training UCM, we use the same parameter settings. We use AdamW optimizer with learning rate 5e-5 and batch size 256. Each time we train the model for 10 epochs. We use warm-up for the first 10\% of iterations. We also use fp16 mix precision to speed up training. 

\begin{table*}[t!]
\large
\centering
\caption{Comparison with other vision-language pre-training models on VQAv2, GQA, NLVR2 and COCO Caption. Our model could achieve competitive or better performance among all models given fewer training images. Evaluation Metrics: for VQA, GQA, NLVR2, results are presented based on the accuracy. For COCO Caption, we follow the common standards to compare the BLEU (B@4), METEOR (R), CIDEr (C) and SPICE (S) scores.}
\resizebox{2\columnwidth}{!}{
\begin{tabular}{l l  c c c c c  c  }
\hline
\multicolumn{2}{c}{\multirow{2}{*}{Tasks}}  & \multirow{2}{*}{ViLBert~\cite{lu2019vilbert} }& \multirow{2}{*}{LXMERT~\cite{tan2019lxmert}} & \multirow{2}{*}{UNITER-base~\cite{chen2019uniter}} & \multirow{2}{*}{ERNIE-VIL-base~\cite{yu2020ernie}} &\multirow{2}{*}{VLP~\cite{zhou2020unified}} & \multirow{2}{*}{UCM (Ours)}\\
\\
\hline
\hline
\multicolumn{2}{c}{Pretrain Images}  & 3m& 180k& 4.2m & 4.2m &  3m & 480k \\
\hline
\multirow{2}{*}{VQA} & test-dev  & 70.55& 72.42& 72.70 & 72.62 &  70.5& \textbf{72.9} \\
 & test-std   & 70.92 & 72.5 & 72.91 &72.85&  70.7 & \textbf{72.9} \\
\hline
\multirow{2}{*}{GQA} & test-dev & -& 60.00& - & - &  -& \textbf{61.3} \\
 & test-std   & - & 60.30 & - &-&  - & \textbf{61.5} \\
\hline
 \multirow{2}{*}{NLVR2} & dev & -& 74.9 & \textbf{75.85 (77.18)} & -&-&\textbf{75.6}\\
 & test-P  & - & 74.5 & \textbf{75.80 (77.85)} &-& -&\textbf{75.5} \\
 \hline
  \multirow{4}{*}{COCO Caption} & B@4 & -& - & - & - & 36.5& \textbf{37.4} \\
 & M  & - & - & - &-&28.4& \textbf{28.8} \\
  & C  & - & - & - &-& 117.7&\textbf{ 119.4 }\\
   & S  & - & - & - &-& 21.3& 21.2 \\
    \hline
 \multirow{4}{*}{\begin{tabular}[c]{@{}l@{}}COCO Caption \\ (CIDEr Optimization)\end{tabular}} & B@4 & -& - & - & - & 39.5& 39.0 \\
 & M  & - & - & - &-& 29.3& 28.8 \\
  & C  & - & - & - &-& 129.3& \textbf{130.1} \\
   & S  & - & - & - &-&23.2& 22.7 \\
\hline
\\
\end{tabular}
}

\label{tab:results}
\end{table*}

\subsection{Finetuning Settings}
We present our finetuning settings for VQAv2~\cite{goyal2017making}, GQA~\cite{hudson2019gqa}, NLVR2~\cite{suhr2017corpus} and COCO Caption~\cite{chen2015microsoft}.

\subsubsection{VQAv2}
VQAv2 dataset~\cite{goyal2017making} is to answer questions given an image. The answering process is usually formatted as a classification task within all possible answers. In our experiments, the VQA questions are appended with the question condition flag before input to the model. We use the final features at position [CLS] to answer the question. We add a two-layer MLP to the final output of [CLS] and use the feature to perform classification. In ablation experiments, we only use default provided data. In the final experiments, following~\cite{chen2019uniter}, we use extra QA data from Visual Genome for data augmentation.

\subsubsection{GQA}
Similar to VQAv2, for GQA dataset~\cite{hudson2019gqa}, we format the problem as a classification task and use the output feature at position [CLS] to answer the questions. Same as VQA, the GQA questions are appended with the question condition flag before input to the model. In ablation experiments, we only use GQA balanced dataset for training. To further help the model adapt to GQA style of questions, in the final experiment, we follow other works~\cite{li2020oscar} to pretrain the model using GQA full set first and then finetune on GQA balanced dataset. 

\subsubsection{NLVR2}
The natural language for visual reasoning for real dataset~\cite{suhr2017corpus} is to answer if the description is correct given two images. The UCM model processes 1 image and 1 language by default. Therefore, we separate the two images into two question and image pairs and process each pair using our proposed model. After getting the [CLS] features for both of the pairs, we simply concatenate the 2 features and use the concatenated feature to perform a binary classification. We noted that in~\cite{chen2019uniter}, a different finetuning process is proposed. For a fair comparison, we compare the results with the same finetuning setting. In NLVR2 experiments, no condition flag is assigned to the sentence as the NLVR2 data does not belong to any type of the pretrained conditions.

\subsubsection{COCO Caption}
We also finetune our model on generation tasks i.g. COCO caption~\cite{chen2015microsoft} on Karpathy’s split. During finetuning, we use one-directional mask only to train the model. During the generation process, the start token is set to [CLS] and [CND] of COCO captioning type. We first use cross-entropy loss to train the captioning model and then apply CIDEr optimization~\cite{rennie2017self} to further improve performance.

\subsection{Ablation Experiments}
\subsubsection{Step by Step Ablation Studies}
\begin{figure*}[]
\centering
\includegraphics[width=1\textwidth,trim=0 0 0 0,clip]{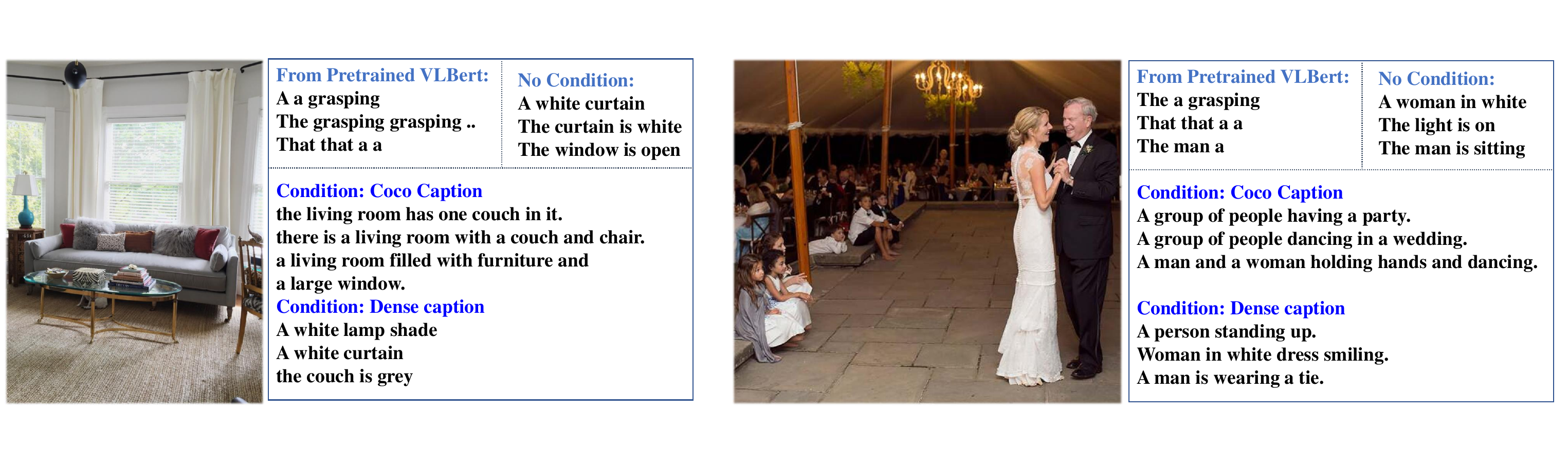}
\caption{An example of generated image descriptions with or without the condition signal. Generation results from a traditional VL-BERT model are not good. For the unified model variations, if the model is not conditional, the generated results are biased to dense captioning style and tend to generate short sentences. Our proposed UCM model could learn a conditional generation language model given different conditional flags.}
\label{fig:5}
\end{figure*}

In this section, we provide step by step ablation studies of our proposed system. The ablation experiments are done on VQAv2, GQA, NLVR2 test-dev set. The results are shown in Table~\ref{table:ablation}. We start by training a baseline model with the same network architecture by only using bi-directional pretraining masks and bi-directional training tasks. We then add 300k images and their original annotations from Conceptual Caption to training data. Results show that simply adding 300k extra image data pairs is unable to improve down-stream finetuning performance much. Moreover, we also try to use LXMERT~\cite{tan2019lxmert} to generate pseudo labels. As pointed out in previous sections and Fig~\ref{fig:5}, the generation quality of LXMERT is not good. Therefore, we observe a huge performance drop when using pseudo labels generated by LXMERT. Furthermore, we perform experiments using UCM with both labeled data and pseudo data generated by the generative model VLP without self-training. Compared with the results only using labeled data, we observe that there is almost no performance improvement. One reason is that the generative model VLP can only generate COCO style captions, therefore the diversity of training data is still limited. Compared with our self-training results, we observe that the self-training method can improve the performance further. After that, we train our proposed UCM model with labeled data only and finetune using the question flag. The result shows simply using UCM and condition flag could improve down-stream finetuning performance. Also, we do one more experiment by removing the condition flag during finetuning. The results drop a little bit if the condition flag is not used. 

We then move to ablation studies of self-training algorithm. We iterate the self-training process by 1 iteration and 2 iterations. We found that based on down-stream finetuning performance, 1 iteration is good enough. Performing 2 iterations is unable to improve performance much.

\begin{figure*}[t]
\begin{center}
\includegraphics[width=0.9\linewidth]{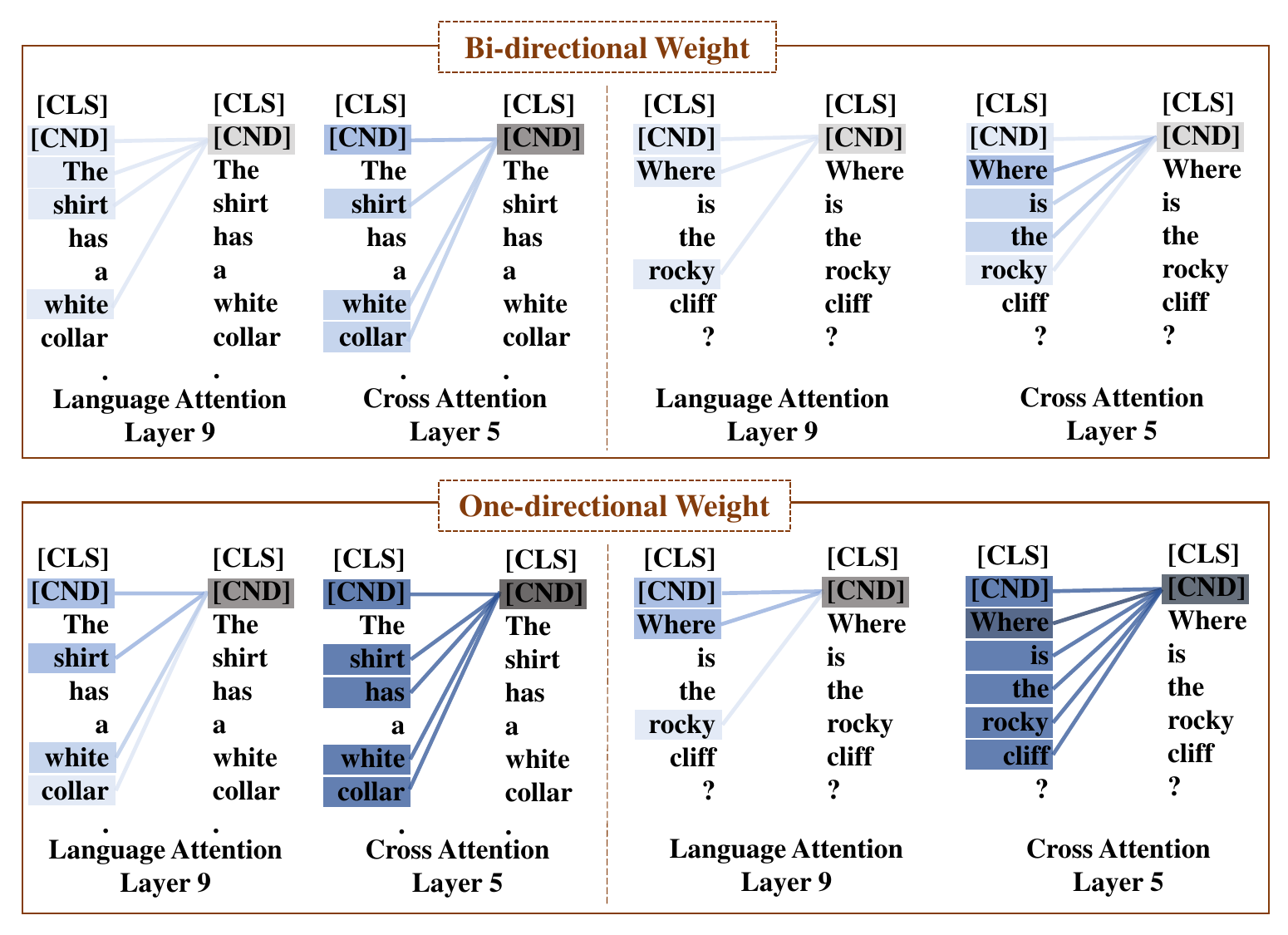}
	\caption{Visualization of the attention map at condition tokens. The darkness of connections represents the attention weight. The darker the color, the higher attention is assigned. Top: Attention masks at bi-directional branch. Bottom: Attention masks at one-directional branch. Left: An example of a dense caption sentence. Right: An example of a question sentence. Based on the visualization results, we can have three conclusions: 1. [CND] mainly affects the one-directional branch. 2. [CND] affects more in deeper layers. 3. [CND] has similar effects on questions and non-questions.}
	\label{fig:7}
\end{center}
\end{figure*}

\begin{table*}[t]
\small 
\centering
\caption{Ablation experiments of our proposed method on VQAv2, GQA and NLVR2.}
\resizebox{0.6\textwidth}{!}{%
\begin{tabular}{llllll}
\hline
\multicolumn{1}{l}{Method} & VQAv2 & GQA & NLVR2  \\ \hline\hline
Baseline & 72.4    & 60.0& 74.9         \\ 
Baseline + Conceptual Caption 300k & 72.4   & 59.9& 75.3   \\ 
UCM with only labeled data & 72.6  & 60.3& 74.8  \\ 
UCM with LXMERT Pseudo Caption & 68.7  & 57.3& 70.1  \\ 
UCM with VLP Pseudo Caption & 72.6  & 60.0& 74.5  \\ 
UCM w/o condition flag & 72.4 & 59.8& 74.8    \\ \hline
UCM + self training step 1  & \textbf{72.8 } & \textbf{60.6}&\textbf{75.6}    \\      
UCM + self training step 2 & 72.7   & 60.5& \textbf{75.6}      \\ 
\hline
\\
\end{tabular}%
}
\\

\label{table:ablation}
\end{table*}

\begin{table*}[]
\small 
\centering
\caption{Additional Ablation Experiments on COCO Caption}
\begin{center}
\resizebox{0.8\textwidth}{!}{%
\begin{tabular}{ccccccccc}
\hline
Model                      & \multicolumn{4}{c}{COCO Caption} & \multicolumn{4}{c}{\begin{tabular}[c]{@{}c@{}}COCO Caption\\ (CIDEr Optimization)\end{tabular}} \\ \hline\hline
                           & B@4    & M      & C      & S     & B@4                    & M                     & C                      & S                     \\ \cline{2-9} 
Baseline                   & 34.9   & 27.1   & 109.4  & 19.9  & 34.0                   & 26.8                  & 117.7                  & 19.6                  \\
UCM with only labeled data & 36.9   & 28.5   & 117.8  & 21.2  & 38.1                   & 28.5                  & 129.9                  & 22.5                  \\
UCM + self training step 1 & 37.4   & 28.8   & 119.4  & 21.2  & 39.0                   & 28.8                  & 130.1                  & 22.7          \\
\hline
\\
\end{tabular}
}

	\label{Albation}
\end{center}
\end{table*}

\begin{figure*}[]

\begin{center}
\resizebox{1\textwidth}{!}{
\includegraphics[]{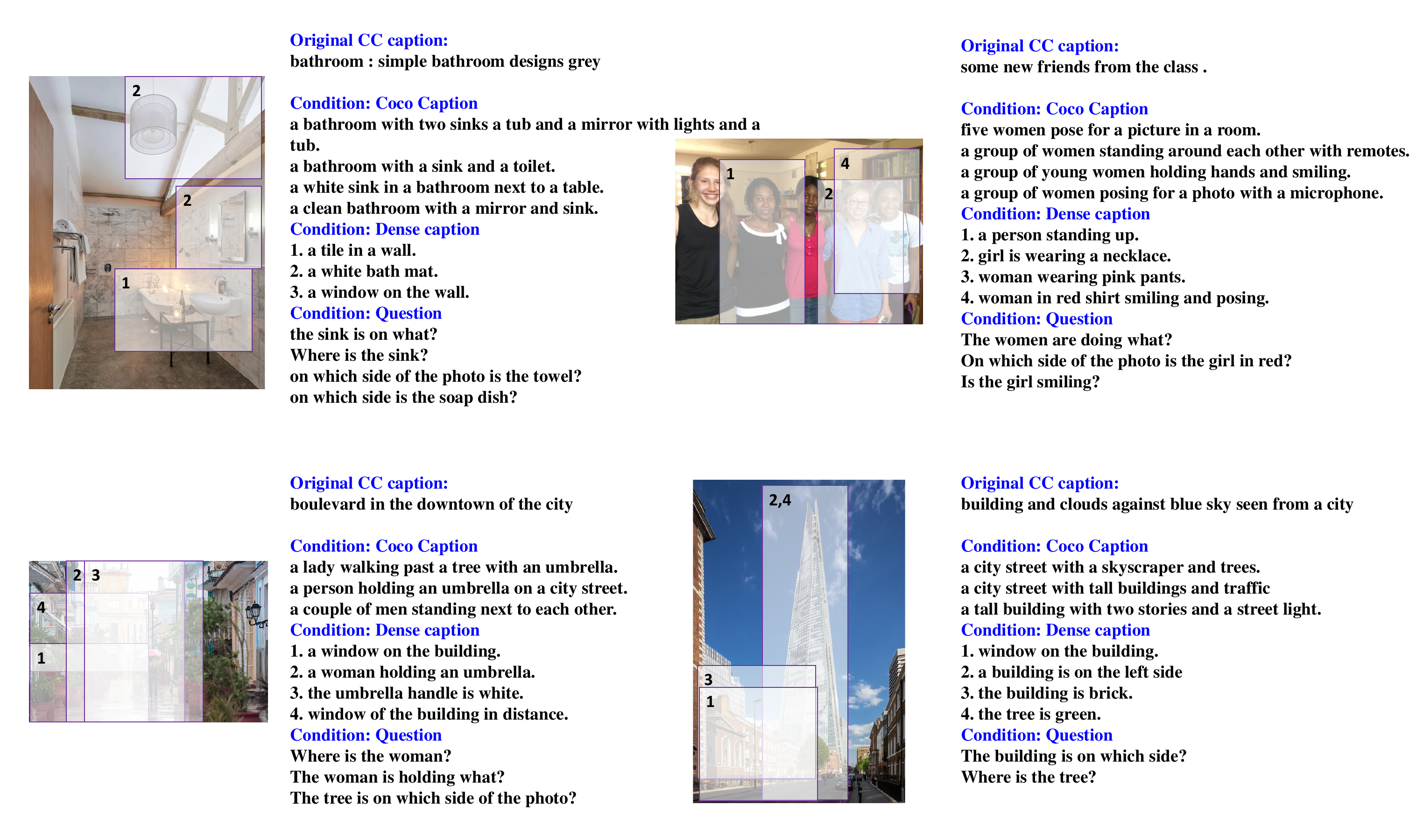}}
	\caption{How the captions are generated given different visual masks (e.g. when some visual regions are masked out). For each generated dense caption, the masked feature region is plotted. Visualization results show that by masking some parts of the image regions, the UCM model could successfully focus on different image areas. }
	\label{generation_results}
\end{center}
\end{figure*}

\subsubsection{Ablation Experiments on Generation Tasks} 

The ablation experiments on generation tasks are shown in Table~\ref{Albation}. The generation process is to predict the next word given the words before and can be formatted as applying a one-directional mask on the sentence tokens. The baseline model gives low performance on COCO captioning task because the model is only trained with bi-directional tasks and is not suitable for generation. We also witness slow convergence speed and training instability. Compared with the baseline model, the UCM model trained with only labeled data can outperform the baseline by a large margin due to the existence of one-directional mask during pretraining. The model also converges faster during experiments. Our UCM model can be finetuned within 40 GPU hours with Nvidia-2080ti GPUs on COCO Caption task. However, the baseline model requires 3 times more finetuning GPU hours. Following the results in previous section, we use self-training step 1 model as the best model. The self-training process also proves its effectiveness in captioning experiments. Compared with the model trained with only labeled data, the self-training model achieves higher accuracy with or without CIDEr optimization. 

\subsubsection{Conditional vs Unconditional}
The experiments in the last section show the effectiveness of the conditional model in finetuning downstream tasks. In this section, we further study how the condition flag affects the generation performance. Results are shown in Figure~\ref{fig:5}. To study this problem, we start by using an online available pretrained vision language BERT model~\cite{tan2019lxmert} to generate captions. Following~\cite{wang2019bert}, we format the generation problem as a sampling problem from Markov Random Field and try to generate languages based on this setting. We found that the generation results are extremely bad using a bi-directional pretrained model. The results are simply repeating several high-frequency words. We then proceed to train a UCM model without using the condition components. We found that the generation results bias to dense captioning styles. This is probably because the training data has much more dense captions than COCO style captions. Finally, we present the results of our UCM model. To further validate the results, we calculate the average generated sentence length. A model trained without condition flag generates sentences with an average length of 4.8 words. Our proposed UCM model can generate diverse image descriptions given different condition flags. When given a condition flag dense caption, the model generates sentences with an average length of 4.7 words. Given a condition flag COCO style caption, our model can generate long sentences with an average length of 10.2 words.

\subsection{Comparison with other methods}
We compare our best UCM model (Self-Training Step1) on VQAv2, GQA, NLVR2 and COCO Caption with other methods. The results are summarised in Table~\ref{tab:results}. We compare our model with similar-sized models (based on BERT base). Our model could achieve competitive or better performance among all models given fewer training images. The VLP model achieves margin advantages compared with our method in COCO Caption (CIDEr Optimization) based on 3 evaluation metrics. The reason is that the model is only pretrained with COCO style captions and  VQA datasets, and no other noisy pseudo captions are included in the pretraining. When the model is used on understanding tasks like VQA, our method prevails with large margins. It proves that our model generalizes better on both generation tasks and understanding tasks.

\subsection{Visualization}

In this section, we give visualizations of the attention map of special tokens and show how the data is generated.

\subsubsection{Understanding the condition token}
We visualize the attention map at condition tokens. As shown in Figure~\ref{fig:7}, we plot the attention weight attending to [CND] position. We plot both the bi-directional branch and the one-directional branch and both a dense captioning style caption and a question. The darker the color, the higher the attention weight. Based on the weights, we could have the following conclusions:

\textbf{[CND] mainly affects the one-directional branch.} We compare the bi-directional weights and one-directional weights (top vs bottom). Although the [CND] flag is used for both branches, the one-directional branch learns to assign higher weights. One reason is that the generation process is more sensitive to the condition flag than the language understanding process. As illustrated in previous sections, our method can generate sentences of different lengths given different condition flags. For the understanding tasks, intuitively the model should focus more on the sentences as a whole. 

\textbf{[CND] affects deeper layers more.} Compared with shallower layers (language attention layer 9), the deeper layers tend to assign higher weights to [CND] position. This is because the deeper layers are more directly related to producing results, thus they rely more on the [CND] flag to control the generation style. 

\textbf{[CND] has similar effects on questions and non-questions.} We compare the visualization of captions and questions (left vs right). No obvious difference can be observed. This implies that the condition flags work in similar ways for caption style sentences and question style sentences.

\subsubsection{Visualization of generation process}
In Figure~\ref{generation_results}, we show how different visual masks affect the language generation. We could have a more obvious observation when generating dense captions. Therefore, for each generated dense caption, we show which feature region is masked. Visualization results show that by masking some parts of the image regions, the UCM model could successfully focus on different image areas and finally produce diverse dense captioning results.  For example, in the first image, when nothing is masked out. The model focuses on the window. When part of the window is masked out, the model will focus on the bath mat and the tile. For COCO style captions, our model also benefits from applying visual masks. Although COCO style captions summarize the whole image, applying visual masks helps the model to look at different areas.

\section{Conclusion and Future Works}
The requirement of paired training data restricts the scale of VL-BERT pretraining. We propose a self-training approach that allows to train VL-BERTs from unlabeled image data. First, we propose UCM -- a vision language BERT that can perform conditional generate directly. Given different condition flags, the unified conditional model can generate dense caption, caption, and even questions. Then we introduce a set of self-training methods for vision language BERT pretraining, including how to generate diverse image descriptions and the self-training pipeline. We also visualize the generation process and the effectiveness of the condition flag. 

For performance, by using the proposed self-training approach and only 300k unlabeled extra data, we are able to get competitive performance within all models with similar model size trained with 3 million extra image data.

\textbf{Future Works.} The use of the conditional model is not restricted to self-training. Future works can be done by exploring more use-cases of the proposed UCM. For example, given an image, our method could be used to generate kid stories, generate advertisement and generate copyright documents with a single pretrained model.

Further extension of training scales could also be explored. Our proposed methods enable training vision language BERTs with unlimited data. One may perform a larger scale of pre-training with more data collected from the Internet. 

\section*{Acknowledgments}
This research is supported by the National Research Foundation, Singapore under its AI Singapore Programme (AISG Award No: AISG-RP-2018-003), the MOE AcRF Tier-1 research grants: RG95/20, and the OPPO research grant.

Fengmao Lv's participation was supported by the National Natural Science Foundation of China (No. 62106204), the Sichuan Natural Science Foundation (No. 2022NSFSC0911, 2022YFG0031), and the Fundamental Research Funds for Central Universities of China (No. 2682022CX068).

\bibliographystyle{IEEEtran}
\bibliography{egbib}

\begin{thebibliography}{10}
\providecommand{\url}[1]{#1}
\csname url@samestyle\endcsname
\providecommand{\newblock}{\relax}
\providecommand{\bibinfo}[2]{#2}
\providecommand{\BIBentrySTDinterwordspacing}{\spaceskip=0pt\relax}
\providecommand{\BIBentryALTinterwordstretchfactor}{4}
\providecommand{\BIBentryALTinterwordspacing}{\spaceskip=\fontdimen2\font plus
\BIBentryALTinterwordstretchfactor\fontdimen3\font minus
  \fontdimen4\font\relax}
\providecommand{\BIBforeignlanguage}[2]{{%
\expandafter\ifx\csname l@#1\endcsname\relax
\typeout{** WARNING: IEEEtran.bst: No hyphenation pattern has been}%
\typeout{** loaded for the language `#1'. Using the pattern for}%
\typeout{** the default language instead.}%
\else
\language=\csname l@#1\endcsname
\fi
#2}}
\providecommand{\BIBdecl}{\relax}
\BIBdecl

\bibitem{devlin2018bert}
J.~D. M.-W.~C. Kenton and L.~K. Toutanova, ``Bert: Pre-training of deep
  bidirectional transformers for language understanding,'' in \emph{Proceedings
  of NAACL-HLT}, 2019, pp. 4171--4186.

\bibitem{lu2019vilbert}
J.~Lu, D.~Batra, D.~Parikh, and S.~Lee, ``Vilbert: Pretraining task-agnostic
  visiolinguistic representations for vision-and-language tasks,'' in
  \emph{Advances in Neural Information Processing Systems}, 2019, pp. 13--23.

\bibitem{wang2019bert}
A.~Wang and K.~Cho, ``Bert has a mouth, and it must speak: Bert as a markov
  random field language model,'' \emph{arXiv preprint arXiv:1902.04094}, 2019.

\bibitem{xie2020self}
Q.~Xie, M.-T. Luong, E.~Hovy, and Q.~V. Le, ``Self-training with noisy student
  improves imagenet classification,'' in \emph{Proceedings of the IEEE/CVF
  Conference on Computer Vision and Pattern Recognition}, 2020, pp.
  10\,687--10\,698.

\bibitem{zoph2020rethinking}
B.~Zoph, G.~Ghiasi, T.-Y. Lin, Y.~Cui, H.~Liu, E.~D. Cubuk, and Q.~Le,
  ``Rethinking pre-training and self-training,'' \emph{Advances in neural
  information processing systems}, vol.~33, pp. 3833--3845, 2020.

\bibitem{brown2020language}
T.~Brown, B.~Mann, N.~Ryder, M.~Subbiah, J.~D. Kaplan, P.~Dhariwal,
  A.~Neelakantan, P.~Shyam, G.~Sastry, A.~Askell \emph{et~al.}, ``Language
  models are few-shot learners,'' \emph{Advances in neural information
  processing systems}, vol.~33, pp. 1877--1901, 2020.

\bibitem{tan2019lxmert}
H.~Tan and M.~Bansal, ``Lxmert: Learning cross-modality encoder representations
  from transformers,'' in \emph{Proceedings of the 2019 Conference on Empirical
  Methods in Natural Language Processing and the 9th International Joint
  Conference on Natural Language Processing (EMNLP-IJCNLP)}, 2019, pp.
  5100--5111.

\bibitem{sharma2018conceptual}
P.~Sharma, N.~Ding, S.~Goodman, and R.~Soricut, ``Conceptual captions: A
  cleaned, hypernymed, image alt-text dataset for automatic image captioning,''
  in \emph{Proceedings of the 56th Annual Meeting of the Association for
  Computational Linguistics (Volume 1: Long Papers)}, 2018, pp. 2556--2565.

\bibitem{singh2020we}
A.~Singh, V.~Goswami, and D.~Parikh, ``Are we pretraining it right? digging
  deeper into visio-linguistic pretraining,'' \emph{arXiv preprint
  arXiv:2004.08744}, 2020.

\bibitem{chen2019uniter}
Y.-C. Chen, L.~Li, L.~Yu, A.~E. Kholy, F.~Ahmed, Z.~Gan, Y.~Cheng, and J.~Liu,
  ``Uniter: Universal image-text representation learning,'' in \emph{ECCV},
  2020.

\bibitem{anderson2018bottom}
P.~Anderson, X.~He, C.~Buehler, D.~Teney, M.~Johnson, S.~Gould, and L.~Zhang,
  ``Bottom-up and top-down attention for image captioning and visual question
  answering,'' in \emph{Proceedings of the IEEE conference on computer vision
  and pattern recognition}, 2018, pp. 6077--6086.

\bibitem{han2019movie}
Y.~Han, B.~Wang, R.~Hong, and F.~Wu, ``Movie question answering via textual
  memory and plot graph,'' \emph{IEEE Transactions on Circuits and Systems for
  Video Technology}, vol.~30, no.~3, pp. 875--887, 2019.

\bibitem{zhang2020action}
J.~Zhang, J.~Shao, R.~Cao, L.~Gao, X.~Xu, and H.~T. Shen, ``Action-centric
  relation transformer network for video question answering,'' \emph{IEEE
  Transactions on Circuits and Systems for Video Technology}, 2020.

\bibitem{guo2021loss}
Y.~Guo, L.~Nie, Z.~Cheng, and Q.~Tian, ``Loss-rescaling vqa: Revisiting
  language prior problem from a class-imbalance view,'' \emph{IEEE Transactions
  on Image Processing}, 2021.

\bibitem{wang2017fvqa}
P.~Wang, Q.~Wu, C.~Shen, A.~Dick, and A.~Van Den~Hengel, ``Fvqa: Fact-based
  visual question answering,'' \emph{IEEE transactions on pattern analysis and
  machine intelligence}, vol.~40, no.~10, pp. 2413--2427, 2017.

\bibitem{guo2021re}
W.~Guo, Y.~Zhang, J.~Yang, and X.~Yuan, ``Re-attention for visual question
  answering,'' \emph{IEEE Transactions on Image Processing}, vol.~30, pp.
  6730--6743, 2021.

\bibitem{yu2018topic}
N.~Yu, X.~Hu, B.~Song, J.~Yang, and J.~Zhang, ``Topic-oriented image captioning
  based on order-embedding,'' \emph{IEEE Transactions on Image Processing},
  vol.~28, no.~6, pp. 2743--2754, 2018.

\bibitem{huang2020image}
Y.~Huang, J.~Chen, W.~Ouyang, W.~Wan, and Y.~Xue, ``Image captioning with
  end-to-end attribute detection and subsequent attributes prediction,''
  \emph{IEEE Transactions on Image Processing}, vol.~29, pp. 4013--4026, 2020.

\bibitem{yan2021task}
C.~Yan, Y.~Hao, L.~Li, J.~Yin, A.~Liu, Z.~Mao, Z.~Chen, and X.~Gao,
  ``Task-adaptive attention for image captioning,'' \emph{IEEE Transactions on
  Circuits and Systems for Video technology}, vol.~32, no.~1, pp. 43--51, 2021.

\bibitem{zhang2020language}
W.~Zhang, C.~Ma, Q.~Wu, and X.~Yang, ``Language-guided navigation via
  cross-modal grounding and alternate adversarial learning,'' \emph{IEEE
  Transactions on Circuits and Systems for Video Technology}, vol.~31, no.~9,
  pp. 3469--3481, 2020.

\bibitem{gao2022efficient}
J.~Gao, X.~Sun, B.~Ghanem, X.~Zhou, and S.~Ge, ``Efficient video grounding with
  which-where reading comprehension,'' \emph{IEEE Transactions on Circuits and
  Systems for Video Technology}, 2022.

\bibitem{vaswani2017attention}
A.~Vaswani, N.~Shazeer, N.~Parmar, J.~Uszkoreit, L.~Jones, A.~N. Gomez,
  {\L}.~Kaiser, and I.~Polosukhin, ``Attention is all you need,'' in
  \emph{Advances in neural information processing systems}, 2017, pp.
  5998--6008.

\bibitem{lu202012}
J.~Lu, V.~Goswami, M.~Rohrbach, D.~Parikh, and S.~Lee, ``12-in-1: Multi-task
  vision and language representation learning,'' in \emph{Proceedings of the
  IEEE/CVF Conference on Computer Vision and Pattern Recognition}, 2020, pp.
  10\,437--10\,446.

\bibitem{li2019visualbert}
L.~H. Li, M.~Yatskar, D.~Yin, C.-J. Hsieh, and K.-W. Chang, ``Visualbert: A
  simple and performant baseline for vision and language,'' \emph{arXiv
  preprint arXiv:1908.03557}, 2019.

\bibitem{li2020unicoder}
G.~Li, N.~Duan, Y.~Fang, M.~Gong, D.~Jiang, and M.~Zhou, ``Unicoder-vl: A
  universal encoder for vision and language by cross-modal pre-training.'' in
  \emph{AAAI}, 2020, pp. 11\,336--11\,344.

\bibitem{li2020oscar}
X.~Li, X.~Yin, C.~Li, P.~Zhang, X.~Hu, L.~Zhang, L.~Wang, H.~Hu, L.~Dong,
  F.~Wei \emph{et~al.}, ``Oscar: Object-semantics aligned pre-training for
  vision-language tasks,'' in \emph{European Conference on Computer
  Vision}.\hskip 1em plus 0.5em minus 0.4em\relax Springer, 2020, pp. 121--137.

\bibitem{zhou2020unified}
L.~Zhou, H.~Palangi, L.~Zhang, H.~Hu, J.~J. Corso, and J.~Gao, ``Unified
  vision-language pre-training for image captioning and vqa.'' in \emph{AAAI},
  2020, pp. 13\,041--13\,049.

\bibitem{yalniz2019billion}
I.~Z. Yalniz, H.~J{\'e}gou, K.~Chen, M.~Paluri, and D.~Mahajan, ``Billion-scale
  semi-supervised learning for image classification,'' \emph{arXiv preprint
  arXiv:1905.00546}, 2019.

\bibitem{sennrich2015improving}
R.~Sennrich, B.~Haddow, and A.~Birch, ``Improving neural machine translation
  models with monolingual data,'' \emph{arXiv preprint arXiv:1511.06709}, 2015.

\bibitem{cheng2019semi}
Y.~Cheng, ``Semi-supervised learning for neural machine translation,'' in
  \emph{Joint Training for Neural Machine Translation}.\hskip 1em plus 0.5em
  minus 0.4em\relax Springer, 2019, pp. 25--40.

\bibitem{wu2019exploiting}
L.~Wu, Y.~Wang, Y.~Xia, Q.~Tao, J.~Lai, and T.-Y. Liu, ``Exploiting monolingual
  data at scale for neural machine translation,'' in \emph{Proceedings of the
  2019 Conference on Empirical Methods in Natural Language Processing and the
  9th International Joint Conference on Natural Language Processing
  (EMNLP-IJCNLP)}, 2019, pp. 4198--4207.

\bibitem{dong2019unified}
L.~Dong, N.~Yang, W.~Wang, F.~Wei, X.~Liu, Y.~Wang, J.~Gao, M.~Zhou, and H.-W.
  Hon, ``Unified language model pre-training for natural language understanding
  and generation,'' \emph{Advances in Neural Information Processing Systems},
  vol.~32, 2019.

\bibitem{radford2019language}
A.~Radford, J.~Wu, R.~Child, D.~Luan, D.~Amodei, and I.~Sutskever, ``Language
  models are unsupervised multitask learners,'' \emph{OpenAI blog}, vol.~1,
  no.~8, p.~9, 2019.

\bibitem{goyal2017making}
Y.~Goyal, T.~Khot, D.~Summers-Stay, D.~Batra, and D.~Parikh, ``Making the v in
  vqa matter: Elevating the role of image understanding in visual question
  answering,'' in \emph{Proceedings of the IEEE Conference on Computer Vision
  and Pattern Recognition}, 2017, pp. 6904--6913.

\bibitem{hudson2019gqa}
D.~A. Hudson and C.~D. Manning, ``Gqa: a new dataset for compositional question
  answering over real-world images,'' \emph{arXiv preprint arXiv:1902.09506},
  vol.~3, no.~8, 2019.

\bibitem{tarvainen2017mean}
A.~Tarvainen and H.~Valpola, ``Mean teachers are better role models:
  Weight-averaged consistency targets improve semi-supervised deep learning
  results,'' \emph{Advances in neural information processing systems}, vol.~30,
  2017.

\bibitem{athiwaratkun2018improving}
B.~Athiwaratkun, M.~Finzi, P.~Izmailov, and A.~G. Wilson, ``Improving
  consistency-based semi-supervised learning with weight averaging,''
  \emph{arXiv preprint arXiv:1806.05594}, vol.~2, no.~9, p.~11, 2018.

\bibitem{ren2015faster}
S.~Ren, K.~He, R.~Girshick, and J.~Sun, ``Faster r-cnn: Towards real-time
  object detection with region proposal networks,'' in \emph{Advances in neural
  information processing systems}, 2015, pp. 91--99.

\bibitem{chen2015microsoft}
X.~Chen, H.~Fang, T.-Y. Lin, R.~Vedantam, S.~Gupta, P.~Doll{\'a}r, and C.~L.
  Zitnick, ``Microsoft coco captions: Data collection and evaluation server,''
  \emph{arXiv preprint arXiv:1504.00325}, 2015.

\bibitem{krishna2017visual}
R.~Krishna, Y.~Zhu, O.~Groth, J.~Johnson, K.~Hata, J.~Kravitz, S.~Chen,
  Y.~Kalantidis, L.-J. Li, D.~A. Shamma \emph{et~al.}, ``Visual genome:
  Connecting language and vision using crowdsourced dense image annotations,''
  \emph{International Journal of Computer Vision}, vol. 123, no.~1, pp. 32--73,
  2017.

\bibitem{yu2020ernie}
F.~Yu, J.~Tang, W.~Yin, Y.~Sun, H.~Tian, H.~Wu, and H.~Wang, ``Ernie-vil:
  Knowledge enhanced vision-language representations through scene graphs,'' in
  \emph{Proceedings of the AAAI Conference on Artificial Intelligence},
  vol.~35, no.~4, 2021, pp. 3208--3216.

\bibitem{suhr2017corpus}
A.~Suhr, M.~Lewis, J.~Yeh, and Y.~Artzi, ``A corpus of natural language for
  visual reasoning,'' in \emph{Proceedings of the 55th Annual Meeting of the
  Association for Computational Linguistics (Volume 2: Short Papers)}, 2017,
  pp. 217--223.

\bibitem{rennie2017self}
S.~J. Rennie, E.~Marcheret, Y.~Mroueh, J.~Ross, and V.~Goel, ``Self-critical
  sequence training for image captioning,'' in \emph{Proceedings of the IEEE
  Conference on Computer Vision and Pattern Recognition}, 2017, pp. 7008--7024.

\end{thebibliography}

\begin{IEEEbiography}[{\includegraphics[width=1in,height=1.25in,clip,keepaspectratio]{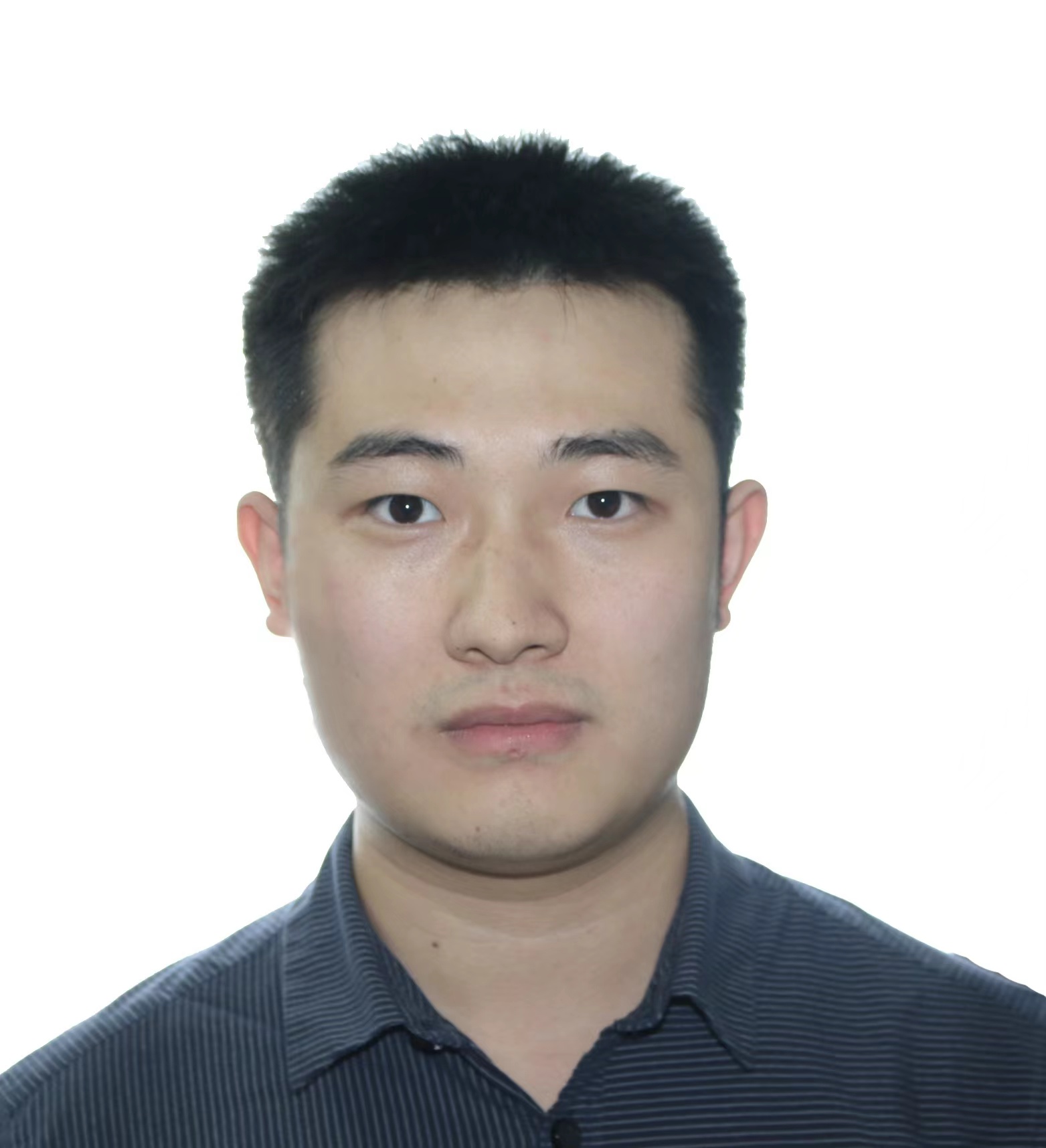}}]{Xiaofeng Yang} is a PhD student at the School of Computer Science and Engineering, Nanyang Technological University, Singapore. His research interests are in computer vision and machine learning.
\end{IEEEbiography}

\begin{IEEEbiography}[{\includegraphics[width=1in,height=1.25in,clip,keepaspectratio]{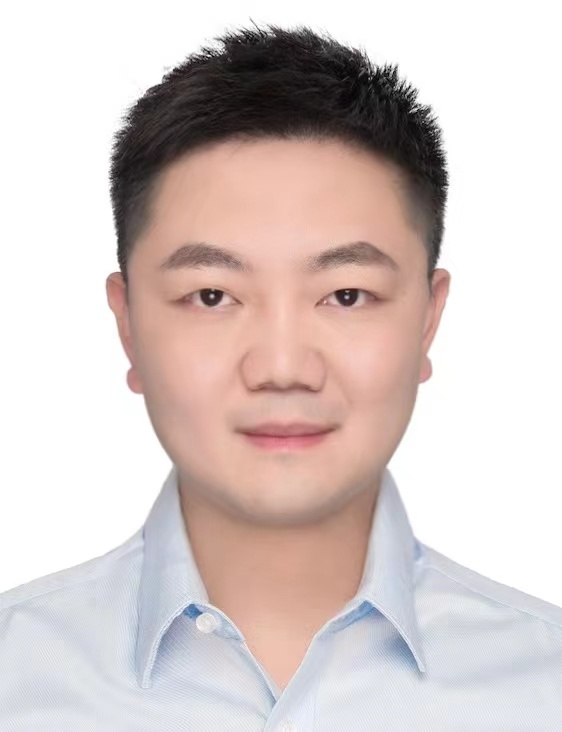}}]{Fengmao Lv} received the bachelor’s and Ph.D. degrees in computer science from the University of Electronic Science and Technology of China, Chengdu, China, in 2013 and 2018, respectively. He is currently an Associate Professor with Southwest Jiaotong University, Chengdu. His research focus includes transfer learning, domain adaptation, and their applications in computer vision and natural language processing.
\end{IEEEbiography}

\begin{IEEEbiography}[{\includegraphics[width=1in,height=1.25in,clip,keepaspectratio]{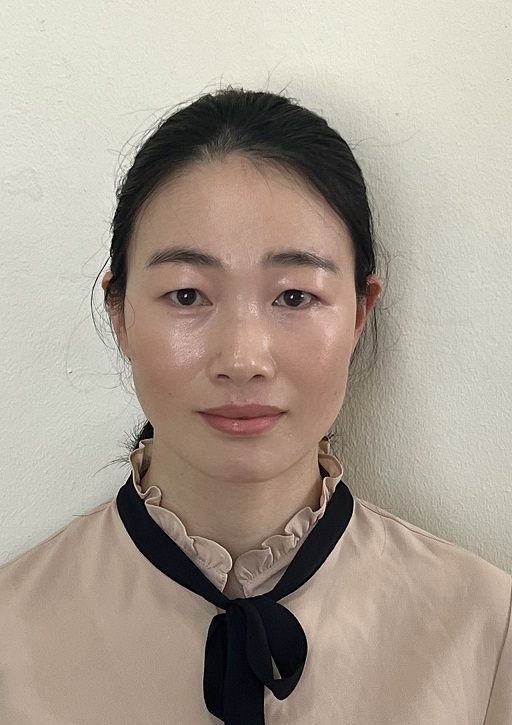}}]{Fayao Liu} is a research scientist at Institute for Infocomm Research (I2R), A*STAR, Singapore. She received her PhD in computer science from the University of Adelaide, Australia in Dec. 2015. Before that, she obtained her B.Eng. and M.Eng. degrees from National University of Defense Technology, China in 2008 and 2010 respectively. She mainly works on machine learning and computer vision problems, with particular interests in self-supervised learning, few-shot learning and generative models. She is serving as an associate editor for IEEE Transactions on Circuits and Systems for Video Technology (TCSVT).
\end{IEEEbiography}

\begin{IEEEbiography}[{\includegraphics[width=1in,height=1.25in,clip,keepaspectratio]{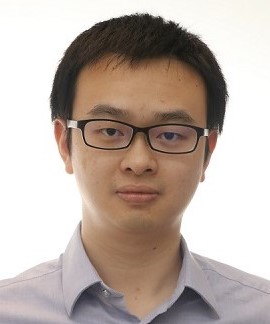}}]{Guosheng Lin} is an Assistant Professor at the School of Computer Science and Engineering, Nanyang Technological University, Singapore. He received his PhD degree from The University of Adelaide in 2014. His research interests are in computer vision and machine learning.
\end{IEEEbiography}

\end{document}